\documentclass[aps,prl,twocolumn,showpacs,superscriptaddress]{revtex4-1}

\usepackage{amsmath,amssymb,amsfonts}
\usepackage{graphicx}
\usepackage{bm}
\usepackage[utf8]{inputenc}
\begin{document}

\title{Additive Multi-Step Markov Chains and the Curse of Dimensionality in Large Language Models}

\author{ O.~V.~Usatenko}
\affiliation{A. Ya. Usikov Institute for Radiophysics and
Electronics Ukrainian Academy of Science, 12 Proskura Street, 61805
Kharkiv, Ukraine\\
Center for Theoretical Physics, Polish Academy of Sciences, Al. Lotnik\'{o}w 32/46, 02-668 Warsaw, Poland
}
%\ead{usatenkoleg@gmail.com}

\author{ S.~S.~Melnyk}
\affiliation{A. Ya. Usikov Institute for Radiophysics and
Electronics Ukrainian Academy of Science, 12 Proskura Street, 61805
Kharkiv, Ukraine}
%\ead{melnik.teor@gmail.com}

\author{ G.~M.~Pritula}
\affiliation{A. Ya. Usikov Institute for Radiophysics and
Electronics Ukrainian Academy of Science, 12 Proskura Street, 61805
Kharkiv, Ukraine}
%\ead{pritula.galina@gmail.com}

\begin{abstract}
Large-scale language models (LLMs) operate in extremely high-dimensional state spaces, where both token embeddings and their hidden representations create complex dependencies that are not easily reduced to classical Markov structures.
In this paper, we explore a theoretically feasible approximation of LLM dynamics using N-order additive Markov chains. Such models allow
the conditional probability of the next token to be decomposed into a superposition
of contributions from multiple historical depths, reducing the combinatorial
explosion typically associated with high-order Markov processes. %The analysis
%focuses on how additive decompositions mitigate some manifestations of the
%curse of dimensionality while simultaneously revealing fundamental limitations arising from non-additive semantic interactions.
The main result of the work is the establishment of a correspondence between an additive multi-step chain and a chain with a step-wise memory function. This equivalence allowed the introduction of the concept of information temperature not only for stepwise but also for additive N-order Markov chains.
\end{abstract}

%\begin{keyword}
%Time series \sep  Random binary sequences  \sep  Additive $N$-order Markov chains   \sep  Memory function  \sep Macroscopic parameters \sep  Information temperature \sep  Large language models
%\end{keyword}
\maketitle

\section{Introduction}
Large Language Models have become the dominant paradigm in contemporary natural language processing~\cite{LLMSurvey}, providing state-of-the-art performance across a wide range of linguistic and reasoning tasks. Despite their empirical success, the internal statistical structure of LLMs remains only partially understood. In particular, it is still unclear which classes of stochastic processes best approximate the sequence distributions generated by LLMs and how these processes relate to classical probabilistic models of text. Developing a mathematically transparent framework that connects LLMs with established sequence models is therefore a pressing and foundational problem (see, in this connection, Refs.~\cite{PhuongHutter,Benechehab} where this point of view is most clearly expressed).  Moreover, current LLM architectures operate as highly complex nonlinear systems whose internal mechanisms are effectively inaccessible to direct mathematical inspection. In this sense, LLMs function as a ``black box,'' and understanding their generative behavior requires the development of new analytical tools beyond classical probabilistic modeling. A rigorous mathematical theory capable of explaining their internal statistical dynamics is therefore urgently needed.

\textbf{Curse of dimensionality and additivity.} A natural starting point for such a framework is the theory of Markov chains. Markovian models have long served as a fundamental tool for describing symbolic sequences, including natural language. Classical $N$-order Markov chains capture dependencies extending up to $N$ previous symbols, but they suffer from a well-known shortcoming: the number of parameters grows exponentially with $N$. As a result, high-order Markov models rapidly become infeasible to estimate or store, even for moderate alphabet sizes. This exponential explosion represents a canonical instance of the \emph{curse of dimensionality} in sequence modeling.

To address these limitations, several alternative models with reduced parameter complexity have been proposed. Among them, additive $N$-order Markov chains constitute a particularly attractive class. In these models, the influence of past symbols decomposes into a sum of contributions associated with different delay positions. Instead of specifying a full transition table of size $O(|A|^N)$, where $|A|$ is the size of token alphabet, an additive chain uses a set of memory functions whose total parameter count grows only linearly with $N$. Additive chains therefore offer a compact yet expressive description of long-range correlations, making them a natural candidate for analyzing systems where high-order dependencies play a central role.

The curse of dimensionality manifests itself in LLMs in multiple ways. High-dimensional token embedding, extremely large parameter spaces, and complex attention interactions~\cite{LLMAttention} all contribute to the enormous computational and statistical cost of training such models. Yet, in contrast to classical high-order Markov chains, LLMs manage to avoid exponential blow-up in memory representation and sampling complexity.
%Understanding the principles that allow LLMs to maintain tractable behavior despite high dimensionality is an important open problem.
Additive Markov chains provide a mathematically transparent setting in which one can investigate how specific structural constraints mitigate dimensionality effects.

These ideas have gained renewed importance in machine learning and artificial
intelligence, where the curse of dimensionality demands compact
macroscopic descriptors of information structure. Among them, the
\textit{temperature} parameter used in LLMs to control
the randomness of generated text plays a role analogous to its thermodynamic
counterpart. The success of LLMs is remarkable precisely because they avoid the curse of dimensionality through architecture and data design.

\textbf{Statistical physics and the curse of dimensionality.}
In statistical physics, quantities such as energy, temperature, chemical potential and pressure summarize the collective behavior of systems with a vast number of microscopic degrees of freedom. A similar principle applies to stochastic processes: although random $N$-order Markov chain which transition conditional probability distribution function (CPDF) depends on numerous microscopic parameters, their large-scale behavior can often be captured by a few effective variables.

The description and understanding of the complexity of natural and artificial dynamical systems remains an open problem in science. The roots of this problem lie at the turn of the 19th and 20th centuries, when statistical mechanics was created. At this time, such new concepts as ergodicity, thermodynamic limit, statistical ensembles and their equivalence, thermodynamic potentials, and many others were born. These roots of the statistical thermodynamics gave rise to new shoots of complexity with the emergence of such new domains of sciences as machine learning, statistical language modeling, artificial intelligence, the study of multidimensional data, and others.

Statistical physics has largely overcome its problems, which led to the construction of statistical thermodynamics. New sciences also find solutions to their problems by formulating new approaches to solving them and finding commonality in their solution. This led to the emergence of a concept that units them - the curse of dimensionality. The curse of dimensionality refers to a set of problems that arise when working with high-dimensional data. The common theme of these problems is that when the dimensionality increases, the volume of the state space increases so fast that the available data become sparse.

Just as statistical physics used new methods of probability theory to solve its problems, the curse of dimensionality problems are solved using the new field of mathematics known as the concentration of measure~\cite{Lafferty}. At the last decades, it became clear that the proper utilization of these phenomena in many area of science might transform the curse of dimensionality into the \emph{blessing of dimensionality}~\cite{Gorban}. In this connection let us remind the main idea of statistical mechanics. If a system can be presented as a union of many weakly interacting subsystems then, in the thermodynamic limit  the whole system can be described by relatively simple deterministic relations in the low-dimensional space of macroscopic variables. It is precisely this idea that we would like to use in application to the Markov chains.

\textbf{Self-attention.} LLMs, although built on deep neural architectures rather than explicit transition matrices, can also be viewed as generative stochastic processes with extremely long memory. Mechanisms such as self-attention~\cite{LLMAttention} allow LLMs to integrate information from distant positions in a sequence, effectively implementing $N$-order Markovian dependence structure of potentially unbounded depth. This raises an intriguing theoretical question: \emph{To what extent can the sequence statistics of LLMs be approximated or interpreted through the framework of additive high-order Markov chains?} Establishing such a connection would not only deepen our understanding of LLM behavior but also reveal new perspectives on the role of dimensionality and long-range dependencies in modern large-scale models.

\textbf{Dichotomy.} A natural first step in the study of additive Markov chains is to restrict attention to binary (dichotomic) sequences. This simplification is justified by several theoretical and methodological considerations. First, the essential complexity of additive Markov models arises from the structure of memory---that is, from the form and strength of delayed influences---rather than from the size of the alphabet. By reducing the alphabet to two symbols, one removes unnecessary combinatorial degrees of freedom and isolates the core mechanism of additivity.

Second, binary sequences admit a particularly transparent statistical description.
When the alphabet is $\{0,1\}$, all correlation functions reduce to
scalar quantities, which makes it possible to derive analytic expressions for
auto-correlations, memory functions, and higher-order dependencies. In contrast,
for alphabets of larger size, the corresponding correlation structure becomes
tensor-valued and significantly more difficult to analyze.

Third, the additive form of the conditional probability takes an especially simple and
tractable shape in the binary setting. See, for example, a particular form of $N$-order additive Markov chain \eqref{CondPr_power}
which leads to a system of linear relations among the parameters. For multi-symbol
alphabets, the analogous expression involves vector- or tensor-valued functions,
resulting in a substantially more complicated parameter space.

Fourth, binary additive chains allow for explicit analytical results that are often
inaccessible in more general settings, including closed-form equations for
correlation functions, exact stationary distributions, explicit forms for the influence of each memory depth, and analytically tractable descriptions of fluctuations and phase-like transitions.

Finally, the binary case is not merely a toy model: many qualitative properties of
additive Markov chains---such as the decay of memory, the structure of
correlations, and the stability of long-range dependence---persist for arbitrary
alphabets. As a result, the binary model serves as a universal minimal framework
from which intuition and theoretical insight can be reliably transferred to
higher-dimensional symbolic systems, including applications related to the analysis
and interpretation of Large Language Models.

\textbf{Macroscopic parameters.} For any statistical theory describing a macroscopic system with many interacting
subsystems to be effective, its dynamics must be represented through a small
number of macroscopic parameters. This fact is confirmed by the entire history of the creation of macroscopic statistical physics.

One of the possibilities for introducing macroscopic parameters is to characterize the complexity of a sequence of random elements. Quantifying the complexity of natural and artificial systems remains a major
scientific challenge~\cite{Bennett,Thurner,Rong,Lavazza,Illiashenko}. Concepts
from statistical mechanics ergodicity, ensembles, and thermodynamic
potentials provide tools for describing the emergence of order and randomness
in high-dimensional systems.

\textbf{Information temperature.} While in LLM temperature governs the variability of generated tokens, in Markov systems information temperature acts as an intrinsic property, arising from the correlation structure of the system or the entropy-energy relationship, and provides a theoretical foundation that links thermodynamic intuition with probabilistic modeling in artificial intelligence.

Here, we further develop the concept of \textit{information temperature}, first introduced in~\cite{UsMPYa} and shown in~\cite{Complexity} to quantify the complexity of random symbolic sequences. By establishing a correspondence between the conditional probability functions of step-wise and $N$-order additive binary Markov chains, we generalize this concept and provide a theoretical rationale for interpreting the temperature parameter in LLMs as a macroscopic measure of informational complexity.

Let us indicate that earlier the idea of usefulness of introducing the concept of information temperature for texts considered as random correlated sequences was proposed in \cite{Mandel}. The concept of ``text temperature'' was applied to linguistic analysis of the texts \cite{Campos,Kosmidis,Rego,Chang} under the assumption that human language could be considered as a physical system. This approach was used afterwords to analyze the  words with the use of the Boltzmann distribution~\cite{Miyazima,Rovenchak}. Note that the use of the Boltzmann distribution to distribute words or tokens in LLM does not have a clear logical justification.

\textbf{The structure }of the rest of the paper is as follows. In Section 2 we provide a brief introduction to the definitions of the key concepts needed for the further derivation of our main results and discussion.
Section 3 contains the main result of the work. It consist in the establishment of a correspondence between an additive multi-step chain and a chain with a step-wise memory function. In Section 4 we present basic results on information temperature. The equivalence between two kind of chains allowed the introduction of the concept of temperature not only for step-wise but also for additive $N$-order Markov chains. Sections 5 and 6 concludes the paper. Here we summarize our results and discuss perspectives for future research.

\section{Background: Basic definitions and concepts}
%%%%%%%%%%%%%%%%%%%%%%%%%%%%%%%%%%%%%%%%%%%%%%%%%%%%%%%%%%%%%%
%
This section provides a brief introduction to the definitions of the key concepts needed for the further discussion. These include concepts such as the symbolic $N$-order Markov chains, the additive $N$-order Markov chains and the Markov chains with step-wise conditional probability distribution function, the equivalence of the $N$-order Markov chain and two-sided random sequences (Ising chains), the Chapman-Kolmogorov equation and the Shannon entropy.

\subsection{Symbolic $N$-order Markov chains and their models}

Consider an infinite random stationary ergodic sequence $\mathbb{S}$  of symbols-numbers $a_{i}$,
\begin{equation}
\label{RanSeq} \mathbb{S}= ..., a_{0}, a_{1},a_{2},...
\end{equation}
taken from the binary alphabet $a_{i}\in \mathcal{A}$:
\begin{equation}\label{alph}
 \mathcal{A}=\{0,1\},\,\, \,\, i \in
\mathbb{Z} = \{...,-1,0,1,2...\}.
\end{equation}
We suppose that the symbolic sequence $\mathbb{S}$ is a \textit{$N$-order Markov chain}. The sequence $\mathbb{S}$ is the $N$-order Markov chain if the probability of symbol~$a_i$ to have a certain value $a\in \mathcal{A}$ under the condition that {\emph{all}} previous symbols are fixed depends only on $N$ previous symbols,
\begin{eqnarray}\label{def_mark}
P(a_i=a|\ldots,a_{i-2},a_{i-1}) = P(a_i=a|a_{i-N},\ldots,a_{i-2},a_{i-1}).
\end{eqnarray}
In this paper we will assume that the CPDF satisfies the conditions,
\begin{eqnarray}\label{def_ergod}
0 < P(a_i= a|a_{i-N},\ldots,a_{i-2},a_{i-1}) <1,
\end{eqnarray}
sufficient to ensure irreducibility and ergodicity.

In works~\cite{UYa}, it  was introduced a model of the $N$-order Markov chain where  by $p_{k}$ the CPDF of symbol ``0'' after an $N$-word with k ones was denoted,
 e.g., after the word
$(\underbrace{11...1}_{k}\;\underbrace{00...0}_{N-k})$, $p_{k}$ is given by
the following expression:
\begin{equation}
p_{k}=P(a_{N+1}=0\mid \underbrace{11\dots
1}_{k}\underbrace{00...0}_{N-k})=\frac{1}{2}+\mu \left(1-\frac{2k}{N}\right).  \label{1}
\end{equation}
Here $N$ is the order of the chain and $\mu$ is the correlation parameter, $-1/2 < \mu <1/2$.

The step-wise chain \eqref{1} biased by the parameter $\nu$ with $p(a)=1/2 + (2a-1)\nu$ (see, e.g., \cite{Biased}) is
\begin{equation}
p_{k} =\frac{1}{2}+\nu+\mu \left(1-\frac{2k}{N}\right).
%\quad k=\{0,1\}, \quad N=1.
\label{p0k}
\end{equation}
In this equation the conditional probability to have a symbol zero after the $N$-word with $k$ number of ones among the preceding symbols contains an additional term $\nu$ describing a difference between the average frequences of ``0''and ``1''.

\subsection{Additive $N$-order Markov chain}

A particular form of the CPDF,
$P(a_i=\alpha^k|a_{i-N},\ldots,a_{i-2},a_{i-1})$, of the binary
$N$-step Markov chain is \cite{MUYa},
\begin{equation}\label{CondPr_power}
  P\left( a_i=1 | a_{i-N}^{i-1} \right)
    =  \overline{a} + \sum_{r=1}^{N} F(r) (a_{i-r}-\overline{a} ),
\end{equation}
where we use the concise notation $a_1^{N}=a_1,a_2,...,a_{N}$. We
refer to such sequences as the \textit{additive} Markov chains and to $F(r)$
as the \emph{memory function} (MF). It describes the strength of
influence of previous symbol $a_{i-r} (1 \leqslant r \leqslant N$)
upon a generated one, $a_{i}$.

Relation between the correlation function,
\begin{equation}\label{KorrDef}
  K(r-r') = \overline{a_{i-r} a_{i-r'}} - \overline{a}^2,
\end{equation}
and the memory functions $F(r)$,
\begin{equation}\label{KorrBin}
K(r)=  \sum_{r'=1}^{N} F(r') K(r-r'), \quad r \geqslant 1,
\end{equation}
allows one to find the memory function as its solution and, thus, to construct a binary
sequence with a prescribed correlation function. The sign $\overline{...}$ in \eqref{KorrDef} means an average; it doesn't matter what kind - ensemble or along a chain - due to the ergodicity, Eq.~\eqref{def_ergod}.

\subsection{Chapman-Kolmogorov equation}
For the binary stationary ergodic Markov chain, the probability $P(a_{1}a_{2}\dots
a_{N})$ of occurring a certain word $(a_{1},a_{2},\dots ,a_{N})$
satisfies the condition of compatibility for the Chapman-Kolmogorov
equation (see, for example, Ref.~\cite{gar}):

\begin{equation}
P(a_{1}\dots a_{N})=\!\!\sum_{a=0,1}\!\!P(aa_{1}\dots a_{N-1})P(a_{N}\mid a,a_{1},\dots
,a_{N-1}).  \label{10}
\end{equation}

\subsection{Markov and two-sided random chains equivalence}
For a \emph{two-sided random chain} equivalent to a binary $N$-order Markov chain the conditional probability that symbol~$a_i$ is equal to unity, under condition that the \emph{rest} of symbols in the chain are fixed, can be presented in the form~\cite{AMUYa},
\begin{equation} \label{2}
P(a_i\!=\!\!1|A_i^-,A_i^+)\!=\!%\displaystyle
\frac{P(a_i=1,A_i^+|A_i^-)}
{P(a_i\!\!=\!\!1,A_i^+|A_i^-)\!+\!P(a_i\!\!=0\!\!,A_i^+|A_i^-)},
\end{equation}
where
$A_i^-=(a_{i-N}\ldots,a_{i-2},a_{i-1})$ and $A_i^+=(a_{i+1},a_{i+2},\ldots,a_{i+N})$
are previous and next words of the length N with respect to symbol $a_i$  .
Here the two-sided conditional probability $P(a_i=1|A_i^-,A_i^+)$ is expressed
by means of the Markov-like probability functions $P(a_i,A_i^+|A_i^-)$.

\subsection{The source entropy}
The Shannon entropy  of subsequence of symbols of length $L$ (see, e.g., Refs.~\cite{Shannon}) is
\begin{eqnarray} \label{entro_block}
H_{L}=-\!\!\!\!\sum_{a_{1},...,a_{L} \in \mathcal{A}} P(a_{1},\ldots,a_{L})\ln
P(a_{1},\ldots,a_{L}).
\end{eqnarray}
Here $P(a_{1},\ldots,a_{L})$ is the probability to
find the $L$-word $(a_{1},\ldots,a_{L})$ in the sequence. Equation~\eqref{entro_block} makes it possible to introduce the
\emph{conditional} entropy, which is the entropy per symbol,
\begin{eqnarray} \label{ShennEntr}
h_{L}= H_{L+1} - H_{L}.
\end{eqnarray}
The
source entropy is the entropy per symbol at the asymptotic limit, $\lim_{L\rightarrow \infty}
 h_L$.  It measures the average information per symbol if all correlations are taken into account.

\section{Macroscopic parameters of the Markov chains
%Temperature of the additive $N$-order Markov chain
}

In this section, we present the main result of our work, namely, the correspondence between step-wise and additive Markov chains.

\subsection{Equivalence between additive and step-wise Markov chains models }

To introduce the notion of the temperature for the random binary additive chain with CPDF \eqref{CondPr_power}, let us rewrite it in the following more general form, representing the probability of generating the symbol $a$, after the word $a_{i-N}^{i-1} $
\begin{equation}\label{CondPr_power1}
  P_{ad}\left( a_i= a | a_{i-N}^{i-1} \right)
    =  p_a + (2a - 1) \sum_{r=1}^{N} F(r) (a_{i-r}-\overline{a} ),
\end{equation}
where $p_1=\overline{a}$ and $p_0=1-\overline{a}$.

We want to establish its correspondence with the chain characterized by CPDF \eqref{p0k} of the chain with step-wise memory
\begin{equation}
P_{sw}(a_i= a |k)=p_{\nu,a} + (2a - 1)\,\, \mu \left(\frac{2k}{N} -1 \right),
\label{p0k1}
\end{equation}
\begin{equation}
p_{\nu,0} =\frac{1}{2} +\nu,\,\,\,\, p_{\nu,1} =\frac{1}{2} - \nu, \,\, k=\sum_{r=1}^{N} a_{i-r} ,
\label{p0k2}
\end{equation}
for which the temperature can be introduced via discussed below Eq.~\eqref{All_tau}. This correspondence can be formulated as a minimization of the ``distance'',
\begin{equation}
Dist = \!\!\!\sum_{a=\{0,1\}}\!\!\! \overline{\left[P_{sw}(a_i= a |k) -   P_{ad}\left( a_i= a | a_{i-N}^{i-1} \right) \right]^2} ,
\label{Dist}
\end{equation}
between the conditional probabilities of two chains from which we should find the parameters $\nu$ and $\mu$ of a sequence with a step-wise memory function. The averaging in Eq.~\eqref{Dist} is performed over the stationary distribution of the initially given additive chain determined by Eq.~\eqref{CondPr_power1}.

The parameters of CPDF in \eqref{CondPr_power1}, $p_0 =1 - p_1$, $p_1 = \overline{a}$ and $F(r)$ are supposed known and the parameters $\mu $ and $\nu$ in \eqref{CondPr_power1} should be determined from the minimization equations. In the absence of correlation, such a connection is obvious:
\begin{equation}
\mu =F =0 \qquad \Rightarrow \qquad p_{\nu,1} = 1/2 - \nu = p_1 = \overline{a} .
\label{MuF}
\end{equation}

If all memory functions in equation \eqref{CondPr_power1} are equal, $F (r) = F_0=Const$, then the additive Markov chain reduces to a Markov chain, Eq.~\eqref{p0k1},  with a step-wise memory function $F_0=2\mu/N$,
\begin{equation}
F (r) = F_0 \qquad \Rightarrow \qquad \mu = N F_0/2 .
\label{f0}
\end{equation}
The result of $Dist$ calculations is
\begin{eqnarray} \label{DistCalc}
  Dist  =   2 h_{0}^{2} + 2\sum_{r,r'=1}^N h_{r}h_{r'} K(r-r'),
\end{eqnarray}
where%
\begin{equation}
  h_{0} =  (\bar{a}-1/2)(1-2\mu) + \nu, \,\,\,\,\,h_{r} = F_{r} - \frac{2\mu}{N}.
\end{equation}

Using Eq.~\eqref{DistCalc} we obtain from minimization equations the following results for the parameters $\mu$ and $\nu$:
\begin{equation}
\frac{\partial Dist}{\partial \mu} = 0, \qquad \Rightarrow \qquad \mu =\frac{ 1 }{2} \frac{ \langle K\star F\rangle }{ \langle\langle K\rangle\rangle  },
\label{Mu}
\end{equation}
\begin{equation}
\frac{\partial Dist}{\partial \nu} = 0, \qquad \Rightarrow \qquad \nu =\frac{1}{2}(1-2\overline{a})(1-2\mu).
\label{Nu}
\end{equation}
Here
\begin{equation}\label{KF}
\langle K\star F\rangle  = \frac{1}{N}\sum_{r,r'=1}^{N} K(r-r')F(r'),
\end{equation}
\begin{equation}\label{AvK}
\langle\langle K \rangle\rangle   =  \frac{ 1}{N^2} \sum_{r,r'=1}^{N} K(r-r') .
\end{equation}

Thus, Eqs.~\eqref{Mu} and \eqref{Nu} define the parameters $\nu$ and $\mu$ of a Markov chain with a step memory function. They are expressed in terms of the microscopic parameters of the additive Markov chain. The parameter $\nu$ has a nearly trivial meaning; it can be determined from a simple comparison of the one-point distribution functions containing in the CPDF \eqref{CondPr_power1} and \eqref{p0k1}. Comparing two results of $\overline{a}$ calculation  with using Eq.~\eqref{CondPr_power1},

\begin{equation}\label{CondPr_power2}
 \overline{a} = p_a =  \overline{P_{ad}\left( a_i= 1| a_{i-N}^{i-1} \right)}, \nonumber
\end{equation}
and with Eq.~\eqref{p0k1},
\begin{equation}
\overline{a} = \overline{P_{sw}(a_i= 1 |k)} = \frac{1}{2} - \nu + \mu (2 \overline{a}  -1) ,\nonumber
\label{p0k3}
\end{equation}
we find $\nu$, Eq.~\eqref{Nu}.

 In a certain sense, the parameter $\mu$ is a more important parameter than $\nu$, since it determines the correlation property of the two chains.
This is what we will discuss in more detail below.

As expected, in the absence of memory $F(r)=0$, we have from Eq.~\eqref{Mu} $\mu=0$.

Let us note another useful form of the parameter $\mu$ presentation with using relation \eqref{KorrBin},
\begin{equation}
 \mu = \frac{1}{2}\frac{\langle K\rangle }{\langle\langle K\rangle\rangle } ,\,\,\,\,
 \langle K \rangle  =  \frac{ 1}{N} \sum_{r=1}^{N} K(r) .
\label{Mu2}
\end{equation}
Despite the apparent similarity of the expressions for $\langle K\rangle$ and $\langle\langle K \rangle\rangle $ in the numerator and denominator of Eq.~\eqref{Mu2}, the sets of correlation functions in them are different; in the numerator, the function represents a set of $K(r)$ with $r=1,2,...,N$, and in the denominator, this is a set of $K(r)$ with $r=0,1,...,N-1$, that provides the important inequality $|\mu|<1/2$  due to the property $|\langle \langle K \rangle\rangle   | > |\langle K \rangle |$.

Another form of equation \eqref{Mu}, expressed in terms of the variance of the random variable $k$,
\begin{equation}\label{Cond_2}
    D(N) = \overline{\left(k - \overline{k} \right)^2} = \sum_{r,r'=1}^{N} K(r-r'),
\end{equation}
is
\begin{equation}\label{mu_centred}
    \mu = \frac{N^2}{2} \frac{\langle K\rangle }{D(N)}.
\end{equation}
In all the above presented expressions, $K(r)$ represents the correlation function of the additive process we are approximating. However, if we assume that the additive and step-wise processes are similar in their correlation properties, then the correlation function of the step process can be used as $K(r)$. Then we can use asymptotic expressions $K(r)$ and $D(N)$ calculated for the step-wise chain in Refs.~\cite{UYa,UYaKM},
\begin{equation}%\label{}
D(N)=\begin{cases} \dfrac{N}{4(1-2\mu )}, & n\gg 1,\\
\dfrac{N^2}{4} -\dfrac{nN(N-1)}{2},& n \ll 1.
\end{cases}
\end{equation}
Here the parameter $n$, defined by
\begin{equation}\label{18a}
n= \frac{N(1-2\mu)}{4\mu},
\end{equation}
determines and separates the strength of persistent ($\mu >0$ ) correlation from weak, $n\gg 1, \mu \rightarrow 0$, to strong, $n \ll 1, \mu \rightarrow 1/2$.

In the case of weak memory and correlation, $|F(r)| \ll 1, \quad |\langle K\rangle| \ll 1$, we can use $D(N) \approx N/4$. Then
\begin{equation}
\mu \approx 2N \langle K\rangle,  \,\,\,\,\, |\mu | \ll 1.
\end{equation}

In the opposite case of the maximal persistence of correlations, the correlator $K(r) \rightarrow 1/4$, $\langle K\rangle\rightarrow 1/4$, and the variance $D(N) \rightarrow N^2/4$. Then it follows from~\eqref{mu_centred} that $\mu \rightarrow 1/2$. Similarly, for the limiting anti-persistent correlations, we have $\langle K\rangle \rightarrow -1/4$,  $D(N) \rightarrow  N^2/4$; therefore, we get $\mu \rightarrow -1/2$.

\section{Basic results on information temperature}
One of the methods for introducing information temperature, which we call the method of equivalent correspondences (EC), is based on the fact that every binary Markov chain is equivalent to some binary two-sided chain \cite{AMUYa} which, in its turn, can be viewed as the Ising sequence for which the probability of a configuration is given by the Boltzmann distribution explicitly containing the temperature $T$.

The second method makes use of the traditional entropy based  thermodynamic definition of temperature with direct calculation of  the block entropy and some \emph{fictive} energy of Markov chain. This method does not suppose a mapping of random sequence on the Boltzmann exponential distribution and describes a broader class of random sequences as compared to the previous EC method.

Both approaches give the same result for the case of nearest neighbor spin/symbol interaction but the method of correspondence of Markov and Ising chains becomes very cumbersome for the chain orders $N \geqslant 3$.
Note that while the first way of the temperature introduction seems quite
natural, the second one can be interpreted as heuristic or axiomatic. We present here  two methods of introducing the information temperature in order to show the validity and effectiveness of the method of introducing \emph{auxiliary, fictitious} energy in the second method.

In both cases, information temperature is understood as an effective thermodynamic quantity that measures the degree of correlation/ordering in a sequence.
%%%%%%%%%%%%%%%%%%%%%%%%%%%%%%%%%%%%%%%%%%%%
\subsection{Temperature from equivalence of the Ising and Markov chains}\label{Equiv}
%%%%%%%%%%%%%%%%%%%%%%%%%%%%%%%%%%%%%%%%%%%
\textbf{N=1}. Using definition \eqref{1} for the CPDF of the ordinary one-step Markov chain, $N=1$, considering the equivalence of Markov chain to the two-sided random sequence \eqref{2} and comparing them with the Boltzmann distribution for the corresponding Ising model,  we get the information temperature as function of parameter
$\mu$,
\begin{equation}
  \mu = \frac{1}{2} \tanh {\left( \frac{1}{\tau} \right)}, \quad\quad
  \frac{1}{\tau} = \frac{\varepsilon}{T} = \frac{1}{2} \ln \frac{1+2\mu}{1-2\mu}.
 \label{MuVsT}
\end{equation}

Under calculations we supposed that the binary variables $a_i=\{0,1\}$ of the Markov chain is related to the spin Ising's variables $s_i=\{-1,1\} = \{\downarrow,\uparrow\}$ by the equality
$s_i = 2 a_i - 1$, and the energy of the spins interaction is of the form: $\varepsilon_{\uparrow
\uparrow }=\varepsilon_{\downarrow\downarrow}=-\varepsilon$,
$\varepsilon_{\uparrow \downarrow
}=\varepsilon_{\downarrow\uparrow}=\varepsilon>0.$
The result \eqref{MuVsT} corresponds to our intuitive understanding the notion of the temperature. Really, the definition Eq.\eqref{1} at $\mu \rightarrow 0$ describes disordered sequences and $\mu \rightarrow 1/2$ does strongly correlated chains. So, for $\tau \rightarrow {\pm\infty}$  we have $\varepsilon/T \simeq 2\mu
\rightarrow 0$, and $\tau \rightarrow \pm 0$ when $\mu \rightarrow
\pm {1/2}$. The negative values of $\tau$ describes an anti-ferromagnetic ordering of spins/symbols $0$ and $1$. We can say that $\tau$ is the temperature $T$
measured in units of  energy $\varepsilon$, $\tau = T/\varepsilon$.

%%%%%%%%%%%%%%%%%%%%%%%%%%%%%%%%%%%%%%%%%%%%
\subsection{Temperature from the entropy}\label{EntrTemp}
%%%%%%%%%%%%%%%%%%%%%%%%%%%%%%%%%%%%%%%%%%%
The second here presented method uses the traditional entropy based thermodynamic definition of temperature with direct calculation of  the block entropy and energy of random  chain in the pair-interaction  approximation.
Unlike the previous method, this approach does not require mapping the sequence onto a Boltzmann distribution and can thus reproduce a broader class of symbolic sequences.
At the beginning, for simplicity, we limit ourselves to the unbiased chain, Eq.~\eqref{1} with $p(a)=1/2$. Here we present only the results of calculations, the details of which can be found in the works \cite{UYa,Complexity}.

Generally, the blocks of words are chosen to be $(N+1)$-length subsequences of symbols occurring  due to the fact that they encompass all the information about the structure of the $N$-step Markov chain. The probabilities of the blocks/words  are defined by the Chapman-Kolmogorov equation \eqref{10}.

\textbf{N=2}. Using Eqs.~\eqref{10} and \eqref{1} for $N=2$, we find the probabilities of 2-symbols words and 3-points probabilities of symbols $``0''$ and $``1''$ occurring.
Introducing some \textit{fictive energies} of the nearest $\varepsilon_1$ and next to nearest $\varepsilon_2$ symbols interactions of different `spin/word' configurations, we calculate the averaged energy, and the entropy per two bonds of random elements. Following the standard procedure of calculating temperature as a derivative of entropy with respect to the  averaged energy, we
obtain the following result:
\begin{equation}\label{2tau}
    \frac{1}{\tau}=\frac{1}{4} \ln
    \frac{1+2\mu}{1-2\mu},
    \end{equation}
where
\begin{equation}\label{AvEps}
    \frac{1}{\tau}=\frac{\langle\, \varepsilon\rangle }{T}, \quad\, \langle\, \varepsilon\rangle=
    \frac{2\,\varepsilon_1+\varepsilon_2}{3}.
\end{equation}
Emerged here average quantity $\langle\, \varepsilon\rangle$, as a result of calculations, can be treated formally as the average (fictive) energy of symbol
interaction.  At the same time, $\langle\, \varepsilon\rangle$ can be considered as
the unit of measurement of the temperature $T$. Putting in Eq.~\eqref{AvEps}
$\varepsilon_1=\varepsilon_2=1$, we obtain the natural (absolute) unit of information temperature measure, $\tau=T$.

Let us note that a surprising point of
the above consideration is independence of Eq.~\eqref{2tau} from the introduced arbitrary energies $\varepsilon_1$ and $\varepsilon_2$ of the interaction of symbols which here are only intermediate tools of temperature normalization.

\textbf{N=3}. Using the same entropy method as presented above we can introduce information temperature for random sequence with the memory length for $N=3$. The result looks more cumbersome than that presented by the equations \eqref{MuVsT} and \eqref{2tau}. The asymptotics of this expression are the following form:
\begin{equation}\label{3N}
    \lim_{\mu \rightarrow 0} \, \left(\frac{1}{\tau}\right) \simeq \frac{2 \mu}{3}, \,\,\,\,\, \lim_{\mu \rightarrow \pm 1/2 }\, \left(\frac{1}{\tau}\right) \propto \ln
    \frac{1+2\mu}{1-2\mu}.
\end{equation}

\textbf{$N \gg 1$}. In Ref.~\cite{UsMPYa} it was shown that for step-wise chain in the high temperature limit -- the large $N$ or small $\mu$ -- the information temperature is

\begin{equation}\label{muN}
   \frac{1}{\tau} = \frac{2 \mu}{N}.
\end{equation}
%
%%%%%%%%%%%%%%%%%%%%%%%%%%%%%%%%%%%%%%%%%%%%
\subsection{Ansatz}
%%%%%%%%%%%%%%%%%%%%%%%%%%%%%%%%%%%%%%%%%%%%
If we compare the  expression
\begin{equation}\label{All_tau}
    \frac{1}{\tau}=\frac{1}{2N} \ln
    \frac{1+2\mu}{1-2\mu}
    \end{equation}
with equations \eqref{MuVsT}, \eqref{2tau}, \eqref{3N} and \eqref{muN}, we can conclude that it coincides with \eqref{MuVsT} at $N=1$, with \eqref{2tau} at $N=2$, asymptotically with \eqref{3N} at $N=3$ and $\mu\rightarrow \pm 1/2$ or $ \mu \rightarrow 0$; and asymptotically with \eqref{muN} at $\mu \rightarrow 0$. This allows us to formulate the following statement:

\textit{Equation~\eqref{All_tau} defines the temperature of the random $N$-order Markov chain with step-wise CPDF, Eq.~\eqref{1}.}

\subsection{The temperature of the additive chain}

Thus, the analysis carried out shows that the parameters $\mu$ and $\nu$  satisfies all the necessary properties and are uniquely determined by the parameters $p_{\nu,a}$ and $F(r)$ of the additive $N$-order Markov chain. Therefore, Eq.~\eqref{All_tau} determines the temperature of additive $N$-order Markov chain.

The reduction of the $N$-order additive Markov chain to its step-wise representation can be viewed as an information-theoretic analogue of statistical averaging in thermodynamics. In statistical physics, microscopic variables describing individual particle states are replaced by their ensemble averages, leading to macroscopic quantities such as temperature and pressure. Similarly, in the coarse-grained approximation of symbolic dynamics, the detailed dependence of the conditional probability on all previous symbols is replaced by an effective macroscopic parameter~$\mu$ that represents the average correlation strength within the sequence. This "information averaging" plays a role analogous to averaging microscopic energies over a physical volume and provides the basis for introducing the concept of \textit{information temperature} as a thermodynamic-like measure characterizing the global degree of order and randomness in the symbolic system.

\begin{figure}[h!]
\begin{centering}
\scalebox{0.5}[0.45]{\includegraphics{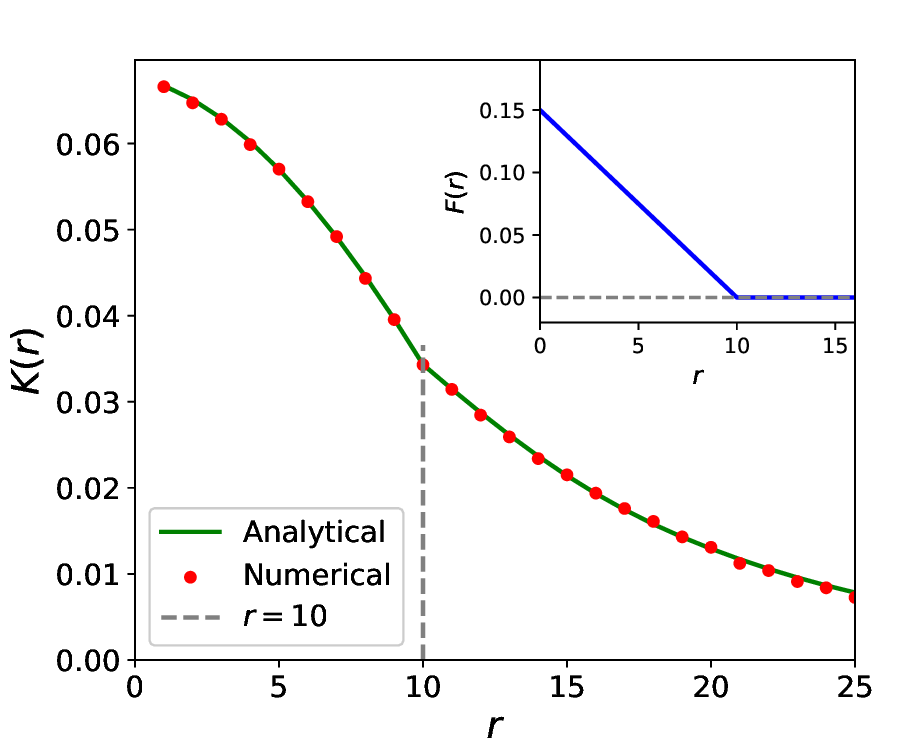}}
\caption{The correlation function $K(r)$ of additive Markov chain constructed using the memory function $F(r)$, Eq.~\eqref{eqmf} (shown in the inset) with memory length $r=N=10$ and parameters $\overline{a}=1/2$ and $F_0=0.15$. The solid line represents the numerical solution of equation~\eqref{KorrBin}. The dots represent the calculations by definition~\eqref{KorrDef} of generating a numerical sequence with CPDF~\eqref{CondPr_power}. } \label{Fig1}
\end{centering}
\end{figure}

\section{Numerical simulations}

This section will illustrate the obtained analytical results. Based on the generation of an additive Markov chain with a linearly decreasing memory function, it will be shown how this chain is put into correspondence with a step-wise Markov chain, and the results for calculating the temperature will be presented.
%In this Section, to verify obtained analytical results, we give examples
%of numerical construction of random sequences with additive CPDF.
It is supposed that the modeled statistical properties of the random
chain are determined by the probability of the symbols $0$ or $1$ occurring,
$p_\alpha =1/2$ and the additive memory function:
\begin{equation}\label{eqmf}
F(r) = \left\{\begin{array}{l}
F_0\left(1 - \frac{r}{N}\right),\qquad   1 \leq r < N, \\[5pt]
0, \quad  \qquad \qquad \qquad r \geq N.
\end{array} \right.
\end{equation}

The results of calculating the correlation function $K(r)$ by two different methods are shown by the solid curve and dots in Fig.~\ref{Fig1}.

\begin{figure}[h!]
\begin{centering}
\scalebox{0.55}[0.45]{\includegraphics{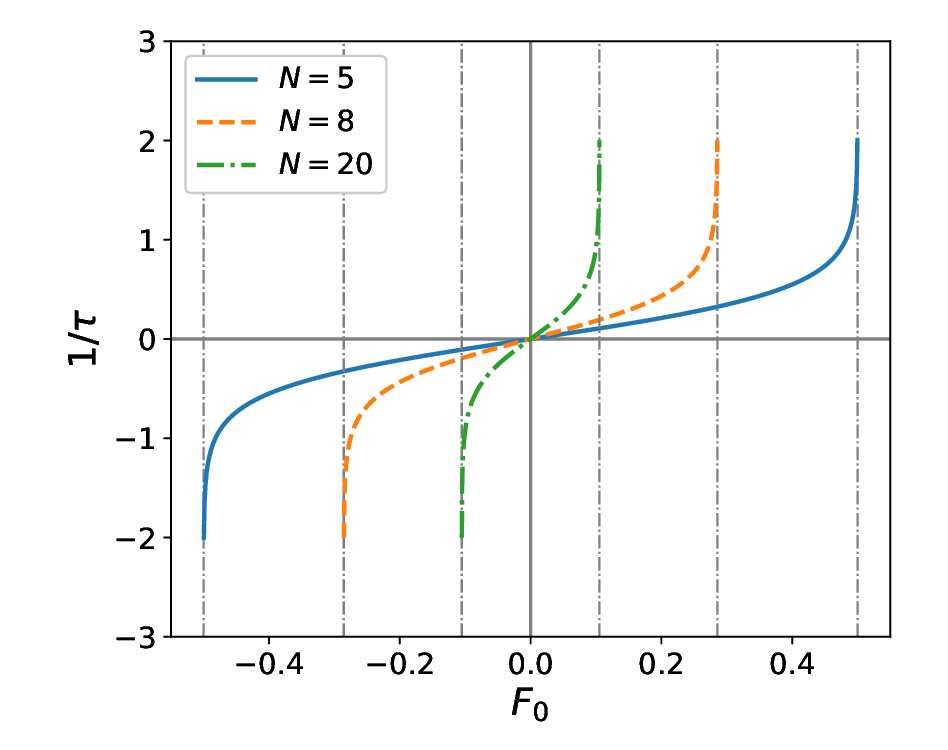}}
\caption{The dependence of inverse temperature $\tau ^{-1}$ defined by Eqs.~\eqref{All_tau} and \eqref{Mu2} for the additive Markov chains with CPDF Eq.~\eqref{CondPr_power1} and memory function~\eqref{eqmf} for $N=5,\,8,\,20$ (the corresponding lines are marked in the legend).  The values of parameter $F_0$ when the  inverse temperature goes asymptotically to infinity are determined by conditions~\eqref{def_ergod}, i.e., $|F_0| \sum_{r=1}^N \left(1 - \dfrac{r}{N}\right)=1.$   } \label{Fig2}
\end{centering}
\end{figure}

Using the calculated values of the correlation function, we find the parameter $\mu$ and then the temperature of the additive Markov chain. The result of these calculations is shown in Fig.~\ref{Fig2} for several values of the parameter $F_0$ of the additive Markov chain memory function.

\begin{figure}[h!]
\begin{centering}
\scalebox{0.45}[0.45]{\includegraphics{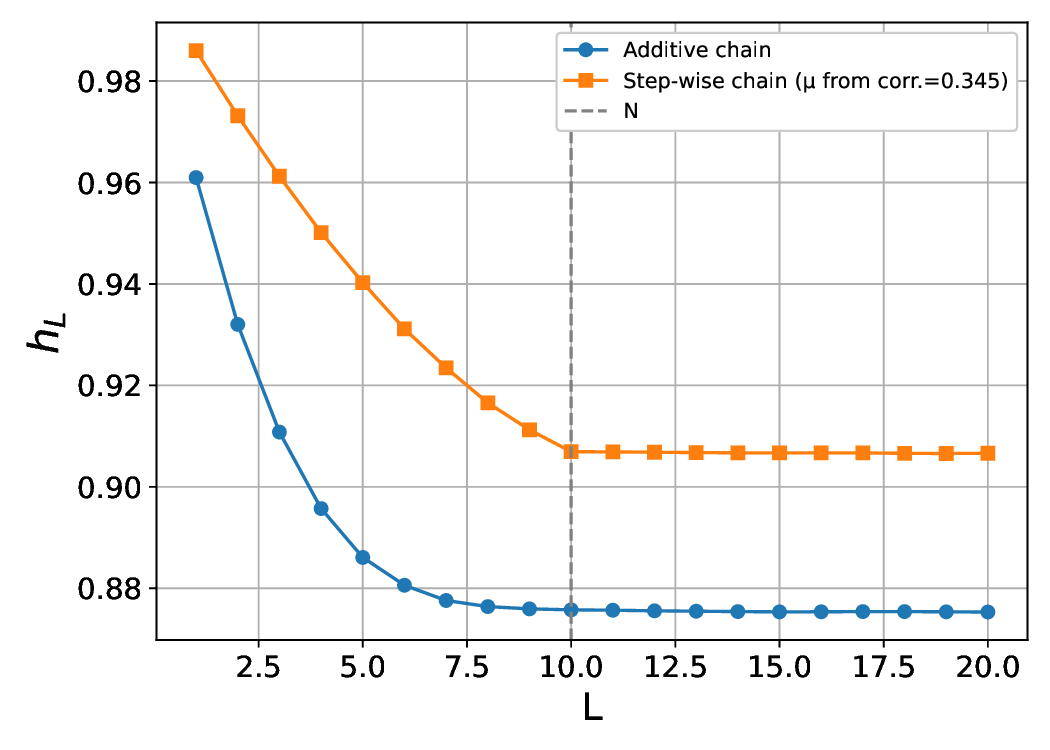}}
\caption{The lower  curve is the dependence of conditional entropies defined by Eqs.~\eqref{entro_block} and~\eqref{ShennEntr}  for the additive Markov chains with CPDF Eq.~\eqref{CondPr_power1} and memory function~\eqref{eqmf} with $N=10$ and $F_0 = 0.15$. The calculated parameter $\mu = 0.345$, defined by the equation~\eqref{Mu2}, gives the entropy of the step-wise chain represented by the upper curve.} \label{Fig3}
\end{centering}
\end{figure}

It is clear that a Markov chain with a step-wise memory function is a coarse-grained description of an additive Markov chain. This coarsening is accompanied by a loss of information, which is reflected in an increase in the source entropy, as follows from Fig.~\ref{Fig3}.

By decreasing the parameter $N$ of the  step-wise Markov chain while maintaining its value for the additive Markov chain and increasing the parameter of correlation $\mu$, we can achieve equality of the entropies of these two chains. This can be viewed as \emph{another principle of equivalence} of random sequences, based on the equality of the source entropies.

\section{Conclusion}

In this work we investigated the statistical structure of random binary
sequences generated by additive $N$-order Markov chains and
established a rigorous correspondence between two distinct
representations of memory in symbolic stochastic processes: the
step-wise conditional probability model and the additive model with a
memory function. This equivalence provides a unified mathematical
framework in which macroscopic characteristics of complex symbolic
systems can be defined and analyzed.

A central outcome of our study is the introduction and justification of
the \emph{information temperature} for additive $N$-order Markov chains.
While the temperature parameter in Large Language Models is
commonly employed as a heuristic tool for adjusting sampling diversity,
its theoretical interpretation has remained unclear. By mapping
high-order additive chains onto step-wise chains, and by exploring both
the equivalence to Ising-type two-sided chains and a thermodynamic
entropy-energy formulation, we demonstrated that an effective
temperature can be consistently defined for a broad class of correlated
binary sequences. This provides a principled basis for understanding
temperature as a macroscopic descriptor of informational complexity.

Our results also reveal a deeper structural analogy between LLM
generation mechanisms and long-memory stochastic processes. Additive
Markov chains avoid the exponential parameter growth of classical
high-order chains, and this property parallels certain architectural
aspects of LLMs that mitigate the curse of dimensionality. The
framework developed here may therefore serve as a conceptual bridge
between symbolic probabilistic models and modern high-dimensional neural
architectures.

Overall, the work provides a theoretical foundation for integrating
concepts from statistical physics, information theory, and stochastic
processes into the study of LLM behavior, opening pathways toward a more
transparent and physically grounded understanding of modern artificial
intelligence systems.

In this work, we have  developed a thermodynamic analogy for symbolic stochastic systems by extending the concept of information temperature to additive binary $N$-order Markov chains. A correspondence between the conditional probability functions of step-wise and additive chains was established, allowing the introduction of a macroscopic parameter that characterizes the balance between order and randomness in symbolic sequences.

The proposed coarse-grained reduction of the high-order Markov model to its step-wise counterpart represents an informational analogue of statistical averaging in thermodynamics, where microscopic fluctuations are replaced by effective macroscopic quantities. In this framework, the parameter~$\mu$ plays the role of an averaged correlation measure, and the information temperature emerges as an intrinsic descriptor of the systems complexity.

The obtained results clarify the conceptual link between thermodynamic temperature and the temperature parameter used in LLMs. Both quantities control the degree of stochasticity in the underlying probability distributions: in physical systems through energy fluctuations, and in symbolic models through information variability. This correspondence provides a theoretical rationale for interpreting the LLM temperature as a macroscopic measure of informational complexity, bridging statistical physics, information theory, and modern generative modeling.

\section{Perspectives}

Several directions for future research emerge naturally from our
findings. Extending the temperature formalism beyond binary alphabets to
multi-symbol systems remains an important challenge, particularly for
applications to natural language. Another promising direction is the
quantitative comparison of empirical LLM-generated sequences with the
predictions of additive $N$-order Markov models, which could clarify the degree to
which LLM statistical behavior can be approximated by low-dimensional
macroscopic parameters. Finally, the temperature-based characterization
of sequence complexity may offer new diagnostic and interpretability
tools for large-scale generative models.

The next step in studying and developing the concept of information temperature and other macroscopic parameters for description is its application to the analysis of symbolic random sequences with an arbitrary finite state space and various memory functions.

The concept of temperature cannot always be introduced into thermodynamics, and even if it is, a system is not always characterized by a single temperature. Similarly, for stationary random sequences, it is important to consider the conditions under which the concept of information temperature can be introduced.

Investigating the informational meaning of the introduced temperature is of interest. In particular, the question arises: can temperature
characterize the academic level of a text or serve
as an indicator of the author's brain activity? The macroscopic interpretation of temperature in information systems could thus serve as a quantitative bridge between physical and artificial intelligence paradigms.

Future work should focus on exploring whether the information temperature can characterize not only structural complexity but also semantic richness in natural language data, or cognitive activity in generative processes. Such studies could deepen our understanding of how thermodynamic principles manifest in symbolic and cognitive systems. These directions may lead to a deeper mathematical theory connecting LLM architectures with interpretable stochastic processes, and may help clarify the fundamental mechanisms driving high-dimensional sequence modeling.

%%%%%%%%%%%%%%%%%%%%%%%%%%%%%%%%%

\textbf{Acknowledgments.}
The authors thanks V.E. Vekslerchik for the discussions during the work.
O.V.U. is grateful to the Center for Theoretical Physics, Polish Academy of Sciences, Warsaw for financial support and hospitality.

%\section*{References}

\end{document}